\documentclass{article}
\usepackage{amsmath,graphicx,mlspconf}
\usepackage{amssymb,amsfonts,amsthm}
\usepackage{xcolor}
\usepackage{algorithmic}
\usepackage{graphicx}
\usepackage{hyperref}
\usepackage{textcomp}
\usepackage[table]{xcolor}
\usepackage{xurl}

\newcommand{\R}{\mathbb{R}}

\newcommand{\matX}{\boldsymbol{X}}

\newcommand{\bb}{\begin{equation}}
\newcommand{\ee}{\end{equation}}
\newcommand{\bbb}{\begin{eqnarray}}
\newcommand{\eee}{\end{eqnarray}}
\newcommand{\benu}{\begin{enumerate}}
\newcommand{\eenu}{\end{enumerate}}

\newcommand{\bpm}{\begin{bmatrix}}
\newcommand{\epm}{\end{bmatrix}}

\newcommand{\ii}{\boldsymbol{i}}
\newcommand{\jj}{\boldsymbol{j}}
\newcommand{\kk}{\boldsymbol{k}}

\newcommand{\quat}[1]{{#1}_0 + {#1}_1 \ii + {#1}_2 \jj + {#1}_3 \kk}

\newcommand{\re}[1]{\mathfrak{Re}\!\left\{#1\right\}}
\newcommand{\im}[2]{\mathfrak{Im}_{#2}\!\left\{#1\right\}}

\newcommand{\hyper}[1]{{#1}_0 + {#1}_1 \boldsymbol{i}_{1} + \dots + {#1}_{d-1}\boldsymbol{i}_{d-1}}

\def\BibTeX{{\rm B\kern-.05em{\sc i\kern-.025em b}\kern-.08em
    T\kern-.1667em\lower.7ex\hbox{E}\kern-.125emX}}

\newtheorem{example}{Example}

\toappear{2026 IEEE International Workshop on Machine Learning for Signal Processing, Sep.\ 28-- Oct.\ 1, 2026, Atlanta, USA}

\title{Ghost Features and Spooky Transfer Learning for Hypercomplex-Valued Neural Networks}
\name{%
   Guilherme Vieira Neto
   \qquad Marcos Eduardo Valle\thanks{Marcos Eduardo Valle acknowledges financial support from the National Council for Scientific and Technological Development (CNPq), Brazil, under grant no 305872/2025-7, and the São Paulo Research Foundation (FAPESP), Brazil, under grant no 2023/03368-0.}%
}
\address{%
   Universidade Estadual de Campinas (UNICAMP), Brazil 
}

\begin{document}
%\ninept

\maketitle

\begin{abstract}
Hypercomplex numbers extend the concept of complex numbers by introducing additional imaginary components. Besides increasing dimensionality, operations on the imaginary parts provide algebraic and geometrical properties that can be beneficial for solving machine learning problems. In this paper, we show how to create hypercomplex-valued neural network layers where the real part corresponds to the output of a traditional real-valued layer. The additional imaginary parts of these hypercomplex-valued layers produce what we call ``ghost features,'' which contain enhanced information that is not present in the output of the real-valued layer. Moreover, ghost features can be effectively integrated into a trained neural network through a process we refer to as ``spooky transfer learning.'' This approach allows us to harness the richness of ghost features, leading to more efficient neural networks. The source code and Jupyter Notebook are available at \href{https://github.com/mevalle/v-nets}{https://github.com/mevalle/v-nets/}.
\end{abstract}

\newcommand{\cem}[1]{\textcolor{blue}{cem: #1}}
\section{Introduction}

Extending real-valued neural networks to hypercomplex domains, including the complex and quaternionic domains, enables the encoding of multidimensional signals while preserving their intrinsic algebraic and geometric structures \cite{Parcollet2020ANetworks,Comminiello2024DemystifyingProcessing,Valle2024UnderstandingProcessing}. Accordingly, hypercomplex-valued neural networks can compactly and consistently represent correlations, phase relationships, and higher-order interactions among input components \cite{hirose12,Grassucci2022PHNNs:Convolutions}. Such representations often enhance generalization and parameter efficiency, particularly in tasks where the underlying data has highly correlated channels or exhibits geometric properties \cite{valous2025computationalalgebras}.

A key feature of hypercomplex-valued networks is the emergence of latent representations arising from algebraic structures computed over observable input variables, which we refer to as ghost features. Ghost features arise from the algebraic constraints of the hypercomplex algebra, including interactions among coupled components, rather than an explicit architectural design of the network. These can be interpreted as internal modes that are not directly recoverable in the network's real-valued projection, equivalent to unobservable components in dynamical systems or hidden states in state-space models. The emergence of ghost features is analogous to the Ghost Module mechanism in GhostNets, wherein redundant or low-cost feature maps are generated from intrinsic representations to reduce computational cost while maintaining representational richness \cite{Han2020GhostNet:Operations}. 
% However, in hypercomplex-valued neural networks, ghost features naturally arise from the internal structure of the representation space rather than an explicit architectural design.

Transfer learning is a machine learning paradigm in which a model trained on a source task is adapted to a related target task \cite{Thrun1998LearningLearn,Geron19HandsOn}. By reusing learned representations, transfer learning can significantly reduce the amount of labeled data and computational resources needed for training in the target domain, accelerate convergence, and enhance generalization.
% —particularly when the target dataset is small or noisy. 
This approach is especially effective when the source and target domains share underlying structural or semantic properties, allowing the model to leverage previously acquired features, invariances, and latent modes. Despite its effectiveness 
% for improving efficiency and performance 
in modern machine learning applications, transfer learning is rarely applied to hypercomplex-valued neural networks due to the scarcity of pre-trained hypercomplex-valued models \cite{Eilers2025InitializingCounterparts}.

This paper formalizes the notion of ghost features in hypercomplex-valued networks, emphasizing how algebra-induced latent structures can be effectively harnessed in pre-training and task adaptation.
% From a theoretical perspective, ghost features in hypercomplex networks can be interpreted as latent modes analogous to unobservable components in dynamical systems or hidden states in state-space models. 
By investigating the behavior of such features under transfer learning, our work introduces the concept of spooky transfer learning as a method of designing hypercomplex-valued neural networks from trained real-valued models.
% In essence, spooky transfer learning captures the highly coupled behavior of ghost features during model transfer.  
% where the term ``spooky'' metaphorically highlights the role of ghost features derived from the network's algebraic structure. 
This paradigm enables pre-trained models to transfer latent structures such as geometric invariances, phase relationships, and higher-order dependencies, to downstream tasks, even when there is weak correspondence between the source and target domains in real-valued terms. 
% By leveraging pre-trained hypercomplex-valued models, it is possible to efficiently reuse these ghost features to enhance learning in new tasks, thereby reducing the amount of data and training required while maintaining structural consistency in the latent space. 
% From a theoretical perspective, ghost features in hypercomplex networks can be interpreted as latent modes analogous to unobservable components in dynamical systems or hidden states in state-space models. 
% Analyzing these features requires tools that extend beyond standard real-valued interpretability techniques, incorporating spectral and geometric methods compatible with the hypercomplex domain.

% In this work, we formalize the notion of ghost features in hypercomplex-valued networks and investigate their behavior under transfer learning. Our study emphasizes how algebra-induced latent structures can be effectively harnessed in pre-training and task adaptation, offering both theoretical insights and practical guidance for designing efficient, geometrically informed neural networks. 

The paper is structured as follows: the next section briefly reviews hypercomplex algebras and hypercomplex-valued neural networks.  Ghost features are introduced in Section \ref{sec:ghost-features} together with a strategy to construct a hypercomplex-valued layer whose real part corresponds to the output of the given real-valued layer. Section \ref{sec:spooky-transfer-learning} shows how ghost features can be incorporated into a pre-trained neural network by means of the so-called ``spooky transfer learning''. An experimental evaluation is conducted in Section \ref{sec:experiments}. The paper finishes with the concluding remarks in Section \ref{sec:conclusion}.

\section{A Brief Review on Hypercomplex-Valued Neural Networks} \label{sec:HvNNs}

% Hypercomplex-valued neural networks are neural networks in which the inputs, outputs, and parameters are represented as hypercomplex numbers. These networks can be understood as traditional neural networks in which hypercomplex operations replace the standard sum and product of real numbers, and hypercomplex-valued activation functions replace conventional real-valued activation functions. 
This section presents a brief overview of the fundamental concepts of hypercomplex algebras and hypercomplex-valued neural networks. For a more detailed exposition, the reader should refer to \cite{Kantor1989HypercomplexAlgebras,Catoni2008TheSpace-Time, Valle2024UniversalNetworks, Zhang2021BeyondParameters}.

\subsection{Hypercomplex Algebra} \label{sub:def}

A hypercomplex number is usually written as
\begin{equation}
\label{eq:hypercomplex-number}
    x = \hyper{x},
\end{equation}
where \(x_0, x_1, \ldots, x_{d-1}\) are real numbers, and \(\ii_1, \ldots, \ii_{d-1}\) are the hyperimaginary units \cite{Kantor1989HypercomplexAlgebras}. The set of all hypercomplex numbers defined by \eqref{eq:hypercomplex-number} is denoted by \(\mathbb{H}\). The component $x_0$ represents the real part of the hypercomplex number $x$ and can be obtained through the mapping $\mathfrak{Re}:\mathbb{H}\to \mathbb{R}$, $\re{x} = x_0$. The other components of a hypercomplex number can be obtained in a similar fashion using the maps $\mathfrak{Im}_{k}:\mathbb{H} \to \mathbb{R}$, where $\im{x}{k}=x_k$, for all $k=1,\ldots,d-1$.

A hypercomplex algebra is a vector space \(\mathbb{H}\) equipped with a bilinear operation that includes a two-sided identity \cite{Valle2024UnderstandingProcessing,Kantor1989HypercomplexAlgebras}. In this framework, we define a hypercomplex number as a linear combination of the elements of the canonical basis \(\tau = \{1, \ii_1, \ii_2, \ldots, \ii_{d-1}\}\), which serves as the basis for \(\mathbb{H}\). The coordinate vector of a hypercomplex number \(x = \hyper{x} \in \mathbb{H}\) with respect to this canonical basis is obtained through the isomorphism \(\varphi: \mathbb{H} \to \mathbb{R}^d\) defined by \cite{Vieira2022AMachines}:
\begin{equation}
\label{eq:isomorphism}
\varphi(x) = [x_0, x_1, \ldots, x_{d-1}] := \vec{x} \in \mathbb{R}^d,
\end{equation}
% To simplify the notation, we will denote $\vec{x}=\varphi(x)$, which is interpreted as a row vector. 
% The set of hypercomplex numbers inherits the metric and the topology of $\mathbb{R}^d$ through the isomorphism $\varphi$. 

A set of hypercomplex numbers inherits the scalar multiplication and the addition from $\mathbb{R}^d$. A hypercomplex algebra is obtained by enriching $\mathbb{H}$ with a multiplication operation, which is 
% the set of hypercomplex numbers with a multiplication operation. 
% The addition of two hypercomplex numbers $x=\hyper{x}$ and $y=\hyper{y}{d-1}$ satisfies 
% \begin{equation}
%     \label{eq:addition}
%     x+y = (x_0+y_0) + (x_1+y_1)\ii_1 + \ldots + (x_{d-1}+y_{d-1})\ii_{d-1}.
% \end{equation}
% Note that $\re{x+y}=x_0+y_0$ and $\im{x+y}{k}=x_k+y_k$.
% The multiplication or product, represented by the symbol “$\times$”, is a bilinear operation that has a two-sided identity. 
% The multiplication or product, 
denoted by “$\times$” and is fully characterized by the products of the basis elements $1,\ii_1,\ldots,\ii_{d-1}$, where the first element $1$ serves as a two-sided identity. Precisely, the product of the hypercomplex units yields hypercomplex numbers 
\begin{equation}
    \label{eq:multiplication-table}
    \ii_\mu \ii_\nu = \hyper{p_{\mu \nu}},
\end{equation}
which form the so-called multiplication table. The multiplication table can be structured as a three-dimensional array of shape $d\times d \times d$ with the coefficients ${p_{\mu\nu}}_k$, for $\mu,\nu,k=0,1,\ldots,d-1$. 
% In particular, we obtain the following matrices by slicing the array in the third component:
% \begin{equation}
% \label{eq:P0}
%     \mathbf{P}_{0} = \begin{bmatrix}
%         1 & 0 & 0 & \ldots & 0 \\
%         0 & {p_{11}}_0 & {p_{12}}_0 & \ldots & {p_{1,d-1}}_0 \\
%         0 & {p_{21}}_0 & {p_{22}}_0 & \ldots & {p_{2,d-1}}_0 \\
%         \vdots & \vdots & \vdots & \ddots & \vdots \\
%         0 & {p_{d-1,1}}_0 & {p_{d-1,2}}_0 & \ldots & {p_{d-1,d-1}}_0 
%     \end{bmatrix},
% \end{equation}
% and 
% \begin{equation}
% \label{eq:Pk}
%     \mathbf{P}_k = \begin{bmatrix}
%         0 & 0 & \ldots & 1 & \ldots & 0 \\
%         0 & {p_{11}}_k & \ldots & {p_{1k}}_k & \ldots & {p_{1,d-1}}_k \\
%         \vdots & \vdots & \ddots & \vdots& \ddots & \vdots \\
%         1 & {p_{k1}}_k & \ldots & {p_{kk}}_k & \ldots & {p_{k,d-1}}_k \\
%                 \vdots & \vdots & \ddots & \vdots& \ddots & \vdots \\
%         0 & {p_{d-1,1}}_k & \ldots & {p_{d-1,k}}_k & \ldots & {p_{d-1,d-1}}_k 
%     \end{bmatrix},
% \end{equation}
% for $k=1,2,\ldots,d-1$. 
Using the matrices $\mathbf{P}_0,\ldots,\mathbf{P}_{d-1}$ obtained by slicing the array in the third component and the isomorphism $\varphi$, the product of two hypercomplex numbers $x$ and $y$ is given by 
\begin{equation}
    \label{eq:multiplication}
    x \times y = \big(\vec{x} \mathbf{P}_0 \vec{y}^T\big) + \sum_{k=1}^{d-1} \big(\vec{x}^T \mathbf{P}_k \vec{y} \big) \ii_k.
\end{equation}
Equivalently, the real and imaginary parts of the product satisfy
\begin{equation}
\label{eq:multiplication-real}
    \re{x \times y} = \vec{x} \mathbf{P}_0 \vec{y}^T,
\end{equation}
and
\begin{equation}
\label{eq:multiplication-imag}
    \im{x \times y}{k} = \vec{x} \mathbf{P}_k \vec{y}^T, \quad \forall k=1,\ldots,d-1.
\end{equation}
We would like to remark that the product of two hypercomplex numbers can also be expressed as a matrix-vector product, which is particularly interesting from a computational perspective \cite{Valle2024UnderstandingProcessing,Grassucci2022PHNNs:Convolutions,Zhang2021BeyondParameters}.

Finally, a hypercomplex algebra is called non-degenerate with respect to the canonical basis $\tau=\{1,\ii_1,\ldots,\ii_{d-1}\}$ if the matrices $\mathbf{P}_0,\ldots,\mathbf{P}_{d-1}$ are all non-singular \cite{Valle2024UniversalNetworks,Vital2022ExtendingNetworks}. Non-degenerate hypercomplex algebras guarantee the universal approximation capability of hypercomplex-valued neural networks and are essential for extracting ghost features.

\begin{example}
\label{ex:quaternions_example}
    Quaternions, developed by Hamilton in 1843 as an extension of complex numbers, are among the most significant and widely used hypercomplex algebras \cite{Parcollet2020ANetworks}. The product of  quaternions is characterized by the matrices 
    \begin{equation*}
        \mathbf{P}_0 = \begin{bmatrix}
            1 & 0 & 0 & 0 \\ 
            0 & -1 & 0 & 0 \\
            0 & 0 & -1 & 0 \\
            0 & 0 & 0 & -1
        \end{bmatrix}, \quad 
        \mathbf{P}_1 = \begin{bmatrix}
            0 & 1 & 0 & 0 \\ 
            1 & 0 & 0 & 0 \\
            0 & 0 & 0 & 1 \\
            0 & 0 & -1 & 0
        \end{bmatrix},
    \end{equation*}
    \begin{equation*}
   \mathbf{P}_2 = \begin{bmatrix}
            0 & 0 & 1 & 0 \\ 
            0 & 0 & 0 & -1 \\
            1 & 0 & 0 & 0 \\
            0 & 1 & 0 & 0
        \end{bmatrix}, \quad \text{and} \quad
        \mathbf{P}_3 = \begin{bmatrix}
            0 & 0 & 0 & 1 \\ 
            0 & 0 & 1 & 0 \\
            0 & -1 & 0 & 0 \\
            1 & 0 & 0 & 0
        \end{bmatrix}, 
    \end{equation*}
    which are derived straightforwardly from the quaternion multiplication table depicted in Table \ref{tab:quaternion}. 
    \begin{table}
        \centering
        \caption{Quaternion multiplication table.}
        \begin{tabular}{c||c|ccc}
  \cellcolor{black!30}$\times$ & \cellcolor{gray!30} $1$ & \cellcolor{red!30}$\ii$ & \cellcolor{green!30}$\jj$ & \cellcolor{blue!30}$\kk$ \\ \hline \hline
 \cellcolor{gray!30}$1$ & \cellcolor{gray!30}$1$ & \cellcolor{red!30}$\ii$ & \cellcolor{green!30}$\jj$ & \cellcolor{blue!30}$\kk$ \\ \hline
 \cellcolor{red!30}$\ii$ & \cellcolor{red!30}$\ii$ & \cellcolor{gray!30}$-1$ & \cellcolor{blue!30}$\kk$ & \cellcolor{green!30}$-\jj$ \\
 \cellcolor{green!30}$\jj$ & \cellcolor{green!30}$\jj$ & \cellcolor{blue!30}$-\kk$ & \cellcolor{gray!30}$-1$ & \cellcolor{red!30}$\ii$ \\
 \cellcolor{blue!30}$\kk$ & \cellcolor{blue!30}$\kk$ & \cellcolor{green!30}$\jj$ & \cellcolor{red!30}$-\ii$ & \cellcolor{gray!30}$-1$
\end{tabular}
        \label{tab:quaternion}
    \end{table}
    For example, the real part of the product of $x = \quat{x}$ and $y = \quat{y}$ satisfies
    \begin{align*}
        \re{x \times y} %&= \vec{x}^T \mathbf{P}_0 \vec{y} \\
        & = \underbrace{\begin{bmatrix}
            x_0, x_1, x_2, x_3
        \end{bmatrix}}_{\vec{x}^T} \underbrace{\begin{bmatrix}
            1 & 0 & 0 & 0 \\
            0 & -1 & 0 & 0 \\
            0 & 0 & -1 & 0 \\
            0 & 0 & 0 & -1
        \end{bmatrix}}_{\mathbf{P}_0}
        \underbrace{\begin{bmatrix}
            y_0 \\ y_1 \\ y_2 \\ y_3
        \end{bmatrix}}_{\vec{y}} \\
        &= x_0 y_0 - x_1 y_1 - x_2 y_2 - x_3 y_3.
    \end{align*}
    while the first imaginary component of the product is given by
    \begin{align*}
        \im{x \times y}{1} &= \vec{x}^T\mathbf{P}_1 \vec{y} 
        &= x_0 y_1 + x_1 y_0 + x_2 y_3 - x_3 y_2.
    \end{align*}
    Since the matrices $\mathbf{P}_0,\ldots,\mathbf{P}_3$ are all non-singular, the quaternions constitute a non-degenerate hypercomplex algebra with respect to the canonical basis $\tau = \{1,\ii,\jj,\kk\}$.
\end{example}

\subsection{Hypercomplex-valued Neural Networks}

Hypercomplex-valued neural networks are analogous to traditional models, but with inputs, outputs, and learning parameters (weights and biases) that are hypercomplex numbers rather than real scalars. The basic operations of addition and multiplication by scalar are performed component-wise, and the multiplication is performed using \eqref{eq:multiplication}. Moreover, 
% the activation function maps hypercomplex numbers into hypercomplex numbers. 
a so-called split activation function $\psi:\mathbb{H}\to \mathbb{H}$ is composed of a real-valued activation function $\psi_{\mathbb{R}}:\mathbb{R} \to \mathbb{R}$ applied component-wise to a hypercomplex number $x = \hyper{x}$ as follows:
\begin{equation}
    \label{eq:split-activation-function}
    \psi(x) = \psi_\mathbb{R}(x_0)+\psi_\mathbb{R}(x_1) \ii_1 + \ldots + \psi_\mathbb{R}(x_{d-1}) \ii_{d-1}.
\end{equation}
% Choosing an appropriate activation function is a topic of research on hypercomplex-valued neural networks. 
The split-ReLU activation function, obtained by \eqref{eq:split-activation-function} with $\psi_{\mathbb{R}} = \mathtt{relu}$, is commonly used in hypercomplex-valued deep learning models \cite{Comminiello2024DemystifyingProcessing,Grassucci2022PHNNs:Convolutions,Zhang2021BeyondParameters,Arena1997MultilayerFunctions}. The following reviews hypercomplex-valued convolutional layers, widely used in deep learning models. 

% A hypercomplex-valued dense layer is a mapping $\mathtt{Dense}(\mathbf{W},\boldsymbol{b}):\mathbb{H}^n \to \mathbb{H}^n$ parametrized by the synaptic weight matrix $\mathbf{W} \in \mathbb{H}^{m \times n}$ and the bias term $\boldsymbol{b} \in \mathbb{H}^m$. A dense layer transforms a hypercomplex-valued input vector $\vetx = [x_1,\ldots,x_n] \in \mathbb{H}^n$ into a hypercomplex-valued output vector $\vety = \mathtt{Dense}(\mathbf{W},\boldsymbol{b})(\vetx)$, with $\vety = [y_1,\ldots,y_m] \in \mathbb{H}^m$ given by the equation
% \begin{equation}
%     \label{eq:dense-layer}
%     y_i = \psi\left( \sum_{j=1}^m x_j \times w_{ji} + b_i \right), \quad \forall i=1,\ldots,m,
% \end{equation}
% where $\psi:\mathbb{H} \to \mathbb{H}$ denotes the activation function. 
% % We would like to point out that a hypercomplex-valued vector $\vetx \in \mathbb{H}^m$ can be written as $\vetx = \hyper{\vetx}$, where $\vetx_0 = \re{\vetx} \in \mathbb{R}^m$ and $\vetx_k = \im{\vetx}{k} \in \mathbb{R}^m$, for $k=1,\ldots,d-1$, are real-valued vectors obtained by taking the real and imaginary parts of $\vetx$.

A hypercomplex-valued convolutional layer is a mapping $\mathtt{Conv2d}(\mathbf{W},\boldsymbol{b}):\mathbb{H}^{H\times W \times C} \to \mathbb{H}^{H' \times W' \times C'}$, where the learnable parameteres $\mathbf{W} \in \mathbb{H}^{P \times Q \times C \times C'}$ and $\boldsymbol{b} \in \mathbb{H}^{C'}$ denote the kernels and the biases, respectively. The pixel value at position \((i, j)\) in the \(k\)th feature channel of a hypercomplex-valued image \(\boldsymbol{X} \in \mathbb{H}^{H \times W \times C}\), where \(H\) is the height, \(W\) is the width, and \(C\) represents the number of hypercomplex-valued channels, is a hypercomplex number denoted as \(x_{ijk} = \hyper{x_{ijk}} \in \mathbb{H}\). This hypercomplex number contains multivariate information, such as color data or spectral signatures for hyperspectral images \cite{Parcollet2020ANetworks,Grassucci2022PHNNs:Convolutions,valous2025computationalalgebras,Gaudet2018DeepNetworks}. 
% Like hypercomplex-valued vectors, a hypercomplex-valued image can be written as $\boldsymbol{X} = \hyper{\boldsymbol{X}}$, where $\boldsymbol{X}_0 = \re{\boldsymbol{X}}$ and $\im{\boldsymbol{X}}{k}$, with $k=1,\ldots,d-1$, are real-valued images of shape $H\times W \times C$ obtained by taking the real and imaginary parts of $\boldsymbol{X}$.
Given a hypercomplex-valued image $\boldsymbol{X} \in \mathbb{H}^{H \times W \times C}$, a hypercomplex-valued convolutional layer yields a hypercomplex-valued image $\boldsymbol{Y} \in \mathbb{H}^{H' \times W' \times C'}$, with $\boldsymbol{Y} = \mathtt{Conv2d}(\mathbf{W},\boldsymbol{b})(\boldsymbol{X})$, whose entries are given by 
\begin{equation}
    \label{eq:hypercomplex-conv}
    y_{ijk} = \psi\left( \sum_{p=0}^{P-1} \sum_{q=0}^{Q-1} \sum_{c=0}^{C-1}  x_{i'j'c} \times w_{pqck} + b_k \right),
\end{equation}
where $i'=i+s p$ and $j'=j+s q$ are translated pixels locations of the image $\boldsymbol{X}$ with stride $s$, and $\psi:\mathbb{H} \to \mathbb{H}$ denotes the activation function. As with traditional convolutional layers, the shape of $\boldsymbol{Y}$ depends on the padding, which is typically either ``same'' or ``valid'' \cite{Geron19HandsOn}.

Like the traditional models, hypercomplex-valued neural networks are universal approximators within a non-degenerate hypercomplex algebra. Specifically, a continuous hypercomplex-valued function \(\boldsymbol{f}:\mathbb{H}^n \to \mathbb{H}\) can be approximated to any desired precision by a hypercomplex-valued MLP network with a single hidden layer, provided it has a sufficient number of neurons and employs a split continuous non-polynomial activation function. Moreover, this holds if the hypercomplex algebra is non-degenerate with respect to the canonical basis \cite{Valle2024UniversalNetworks}. 
The approximation capability holds, in particular, for complex-, quaternion-, and Clifford-valued MLP networks that utilize the split-ReLU activation function \cite{Arena1997MultilayerFunctions,Buchholz2008OnPerceptrons,Vital2022ExtendingNetworks}. Briefly, the resulting hypercomplex-valued MLP model can be derived by applying the approximation theorem of traditional MLP networks to the real and hypercomplex parts of the hypercomplex-valued function \(\boldsymbol{f}:\mathbb{H}^n \to \mathbb{H}\). The concepts of ghost features and spooky transfer learning discussed in the following are based on this latter remark. 

\section{Ghost Features} 
\label{sec:ghost-features}

The first layers of a neural network are responsible for extracting informative, structured features necessary for success in machine learning tasks. 
% In image processing tasks, the first convolutional layers learn low-level features that capture local, simple patterns, such as edges and textures. 
% By capturing these elementary structures, the first layers reduce noise and variability while preserving essential information, thereby facilitating the learning process in subsequent layers. 
As pointed out by Han et al. \cite{Han2020GhostNet:Operations}, ``abundant and even redundant information in the feature maps of well-trained deep neural networks often guarantees a comprehensive understanding of the input data.'' Accordingly, in computer vision tasks, similar pairs of feature maps may exist in which one appears to be a ghost of the other. Moreover, the ghost maps can be obtained via simple linear operations and, when coupled with a convolutional layer, can reduce the network's parameter count and computational complexity \cite{Han2020GhostNet:Operations}. Similarly, by appropriately interpreting a real-valued input as hypercomplex numbers, one can embed a traditional layer within a hypercomplex layer, where the imaginary parts of the output represent ghost features generated by simple linear operations while the real part replicates the original output features. The following details this remark for hypercomplex-valued convolutional layers with split activation function.
% In the following, we detail this remark for dense and convolutional layers with split activation functions. 
% However, instead of beginning with a real-valued layer, we will start with a hypercomplex-valued layer and show that its real part corresponds to a traditional layer. 

Let $\mathbb{H}$ be a hypercomplex algebra with dimension $d$. From \eqref{eq:multiplication-real} and \eqref{eq:hypercomplex-conv}, the real part of the $ijk$th entry of the output $\boldsymbol{Y}=\mathtt{Conv2d}(\mathbf{W},\boldsymbol{b})(\boldsymbol{X})$ of a convolutional layer with split activation function satisfies
\begin{align*}
    &\re{y_{ijk}} 
    = \psi_{\mathbb{R}}\left( \sum_{p,q,c} \re{x_{i'j'c} \times w_{pqck}} + \re{b_k} \right) \\
    &= \psi_{\mathbb{R}}\left( \sum_{p,q,c} \vec{x}_{i'j'c}  \mathbf{P}_0 \vec{w}_{pqck} + {b_k}_0 \right) \\
    &= \psi_{\mathbb{R}}\left( \sum_{p,q,c} \sum_{\mu=0}^{d-1} x^{(\mathbb{R})}_{i'j',d(c-1)+\mu}  w^{(\mathbb{R})}_{pq,d(c-1)+\mu, k} + {b_k}_0 \right)
\end{align*}
where $x^{(\mathbb{R})}_{i'j',d(c-1)+\mu}$ are the entries of a real-valued image $\boldsymbol{X}^{(\mathbb{R})} \in \mathbb{R}^{H \times W \times (dC)}$ obtained by concatenating the real and imaginary parts of the hypercomplex-valued feature channels: 
\begin{equation}
    \label{eq:concatenate_channels}
    \boldsymbol{X}^{(\mathbb{R})} =[\underbrace{{\boldsymbol{X}_1}_0,\ldots,{\boldsymbol{X}_1}_{d-1}}_{\substack{\text{real and imag. parts} \\ \text{of channel \#1}}},\ldots,\underbrace{{\boldsymbol{X}_C}_0,\ldots,{\boldsymbol{X}_C}_{d-1}}_{\substack{\text{real and imag. parts} \\ \text{of channel \#C}}}].
\end{equation}
Similarly, $w^{(\mathbb{R})}_{pq,d(c-1)+\mu, k}= \sum_{\nu=0}^{d-1} {p_{\mu\nu}}_0 {w_{pqck}}_\nu$ are the entries of a real-valued convolution filter $\boldsymbol{W}^{(\mathbb{R})} \in \mathbb{R}^{P \times Q \times (dC) \times C'}$. Equivalently, using colon notation, we have 
\begin{equation}
    \label{eq:real_filter_weights}
    {\boldsymbol{w}^{(\mathbb{R})}_{pq:k}} = [\mathbf{P}_0 \vec{w}_{pq1k},\ldots,\mathbf{P}_0\vec{w}_{pqCk}] \in \mathbb{R}^{dC}.
\end{equation}
From the last identities, we conclude that
\begin{equation}
    \label{eq:conv2real}
    \re{\mathtt{Conv2D}(\mathbf{W},\boldsymbol{b})(\boldsymbol{X})} = \mathtt{Conv2D}_{\mathbb{R}}\left(\mathbf{W}^{(\mathbb{R})},\boldsymbol{b}^{(\mathbb{R})}\right)\left(\boldsymbol{X}^{(\mathbb{R})}\right),
\end{equation}
where $\mathtt{Conv2D}_{\mathbb{R}}(\mathbf{W},\boldsymbol{b}):\mathbb{R}^{W \times H \times (dC)} \to \mathbb{R}^{W' \times H' \times C'}$ is a traditional real-valued convolutional layer applied to the image with the hypercomplex-valued components of the features concatenated. The weights of this layer are given by \eqref{eq:real_filter_weights} and the bias vector is given by $\boldsymbol{b}^{(\mathbb{R})} =\re{\boldsymbol{b}}$, with $\re{\cdot}$ applied componentwise. 

Conversely, given a traditional real-valued convolutional layer with kernel $\mathbf{W}^{(\mathbb{R})} \in \mathbb{R}^{P \times Q \times (dC) \times C'}$ and a bias term $\boldsymbol{b}^{(\mathbb{R})}$, the hypercomplex-valued convolutional layer with kernel $\mathbf{W} \in \mathbb{H}^{P \times Q \times C \times C'}$ given by 
\begin{equation}
    \label{eq:real2hypercomplex_conv_weights}
    w_{pqck} = \varphi^{-1}\left(\mathbf{P}_0^{-1} \begin{bmatrix}
        w^{(\mathbb{R})}_{ij,d(c-1)+1,k} \\ \vdots \\ w^{(\mathbb{R})}_{ij,dc,k}
    \end{bmatrix}  \right) \in \mathbb{H},
\end{equation}
and bias $\boldsymbol{b} \in \mathbb{H}^{C'}$ such that $\re{b_k}=b_k^{(\mathbb{R})}$ for all $k=1,\ldots,C'$, satisfies \eqref{eq:conv2real}, with the hypercomplex-valued image $\boldsymbol{X} \in \mathbb{H}^{H\times W \times C}$ obtained from the input $\boldsymbol{X}^{(\mathbb{R})} \in \mathbb{R}^{H\times W \times (dC)}$ using the inverse isomorphism $\varphi^{-1}$. Furthermore, the imaginary parts 
\begin{equation}
    \label{eq:ghost-conv}
    \im{\mathtt{Conv2d}(\mathbf{W},\boldsymbol{b})(\boldsymbol{X})}{k} \in \mathbb{R}^{H' \times W' \times C'},
\end{equation}
are ghost features of the hypercomplex-valued convolutional layer for all $k=1,\ldots,d-1$. The following example illustrates some ghost feature maps of the first convolutional layer of the EfficientNetV2 (B0) pre-trained on ImageNet.
% regarded for great performance across multiple image domains.

\begin{example}
\label{ex:effnet_example}
EfficientNets, introduced by Tan and Le in 2019 and further improved in 2021 \cite{tan2021efficientnetV2}, are highly compact computer vision models designed via a compound scaling factor to uniformly scale network depth, width, and resolution. We considered an EfficientNetV2 (B0) pre-trained on ImageNet and replaced the first real-valued convolutional layer with a quaternion-valued layer, ensuring that \eqref{eq:conv2real} holds. Figure \ref{fig:cat-ghost-features} shows an input color image and a few features extracted by the first hypercomplex-valued convolutional layer. Precisely, the first column shows the real-valued features equivalent to features yielded by the traditional pre-trained model, while the following columns correspond to the respective ghost features obtained using \eqref{eq:ghost-conv}. 
% Note that the ghost features associated with the imaginary components comprise information that can be as useful as the real-valued features for a computer vision task. We confirm this claim in the following section by proposing the spooky transfer learning.

\begin{figure}[t]
    \centering
    \scriptsize{\textsf{Original image}}\\
    \includegraphics[width=0.2\linewidth]{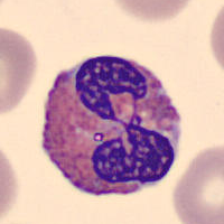}\\
    \includegraphics[width=\linewidth]{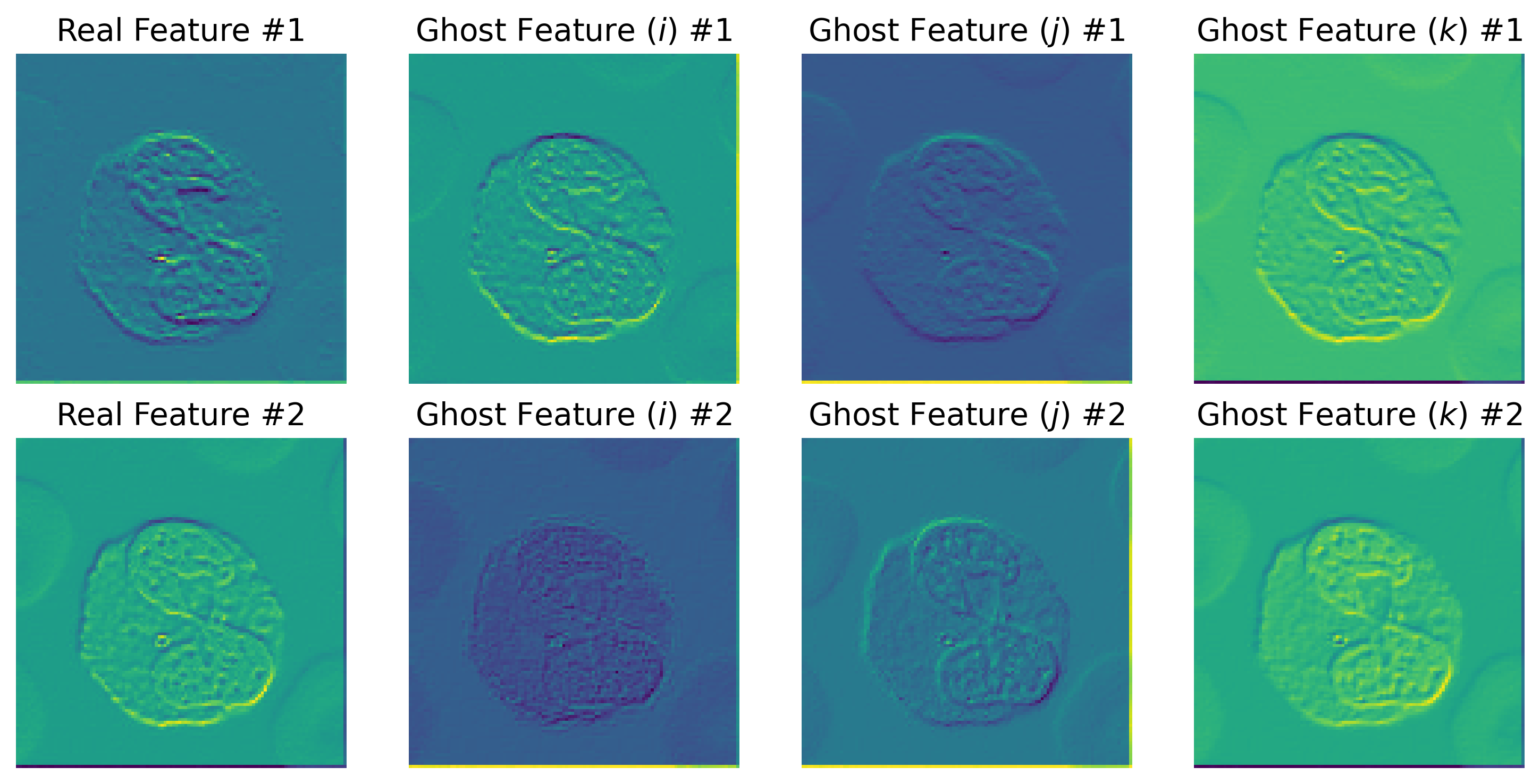}
    \caption{Image from BloodMNIST \cite{medmnistv2} followed by real and ghost features obtained from the first convolutional layer of a pre-trained EfficientNetV2 (B0) embedded into quaternions.}
    \label{fig:cat-ghost-features}
\end{figure}
\end{example}

\section{Spooky Transfer Learning}
\label{sec:spooky-transfer-learning}

Transfer learning consists of adapting the feature-extraction and processing structures of a pre-trained network for a new model, thereby leveraging previously trained feature maps to enhance downstream task performance \cite{Thrun1998LearningLearn,Geron19HandsOn}. This paradigm is widely adopted in deep learning applications; however, the limited availability of pre-trained hypercomplex-valued models makes it difficult to effectively leverage transfer learning in hypercomplex domains. 
% In this context, spooky transfer learning is a method for deriving hypercomplex-valued models from pre-trained real-valued ones, producing feature maps that contain the original feature extractors 
% as well as additional ghost features arising from the underlying algebraic structure. 
% This section details spooky transfer learning and showcases its application on a benchmark dataset, comparing it to traditional transfer learning.

The concept of spooky transfer learning involves embedding a real-valued layer from a pre-trained neural network into a hypercomplex-valued layer.
% , followed by a ``dimension reduction'' layer. 
% Accordingly, the output from the hypercomplex layer retains the original features from the real-valued layer in its real part, along with ghost features in its imaginary part. 
This embedding increases the total number of features, which is mitigated by a dimension reduction layer that selects the most relevant features for the machine learning task. Formally, let $\mathcal{L}_\R(\mathbf{W}^{(\R)},\mathbf{b}^{(\R)})$ denote a layer in a pre-trained deep learning model with parameters $\mathbf{W}^{(\R)}$ and $\mathbf{b}^{(\mathbb{R})}$, and let $\mathbb{H}$ be a non-degenerate $d$-dimensional hypercomplex algebra. In analogy to \eqref{eq:conv2real}, one obtains hypercomplex-valued parameters $\mathbf{W}$ and $\mathbf{b}$ such that:
\begin{equation}
    \label{eq:conv2real_sec4}
    \re{\mathcal{L}(\mathbf{W},\mathbf{b})(\matX)} = \mathcal{L}_\R(\mathbf{W}^{(\R)},\mathbf{b}^{(\R)})(\matX^{(\R)}),
\end{equation}
where the input signal $\matX$ is obtained from $\matX^{(\R)}$ using $\varphi^{-1}$. From \eqref{eq:conv2real_sec4}, the real part of the hypercomplex-valued layer matches the feature generated by the original layer. Additionally, the $d-1$ imaginary parts of $\mathcal{L}(\mathbf{W},\mathbf{b})(\matX)$ are ghost features yielded by the underlying structure of the product in $\mathbb{H}$. Notably, this output contains $d$ times as many channels as the original layer's output. In order to mesh the output correctly into the original pipeline, we add a trainable dimension reduction layer, in the form of a dense layer if the original layer was dense, or a convolutional layer with $1 \times 1$ kernels if the original layer was convolutional. It is important to highlight that $\mathbf{W}$ and $\mathbf{b}$ are derived from the pre-trained network and are frozen during training, as in traditional transfer learning. In other words, the pre-trained layer produces a non-trainable hypercomplex-valued layer that yields additional ghost features, and a subsequent dimension-reduction layer serves as a trainable layer that combines the original features with the ghost features, thereby enhancing feature extraction.

\section{Experimental Evaluation}
\label{sec:experiments}

To showcase the enhanced feature extraction and predictive power achieved by spooky transfer learning, we apply the proposed pipeline to an EfficientNetV2 (B0) \cite{tan2021efficientnetV2} for the classification task in the BloodMNIST dataset. MedMNIST-v2 is a large-scale biomedical image dataset for classification tasks \cite{medmnistv2} containing pre-processed images from several medical specialty domains. BloodMNIST is one such set featuring over 17,000 blood cell microscope images organized into 8 classes, with a provided train-validation-test split of 7:1:2. This dataset was selected as a suitable experiment due to being readily available, broadly documented, and complex enough to warrant the use of deep learning models. Moreover, it is particular to the medical domain, emphasizing the importance of feature extraction and the ability to adapt existing pretrained models by leveraging their representational power.

\begin{figure}
    \centering
    \includegraphics[width=\columnwidth]{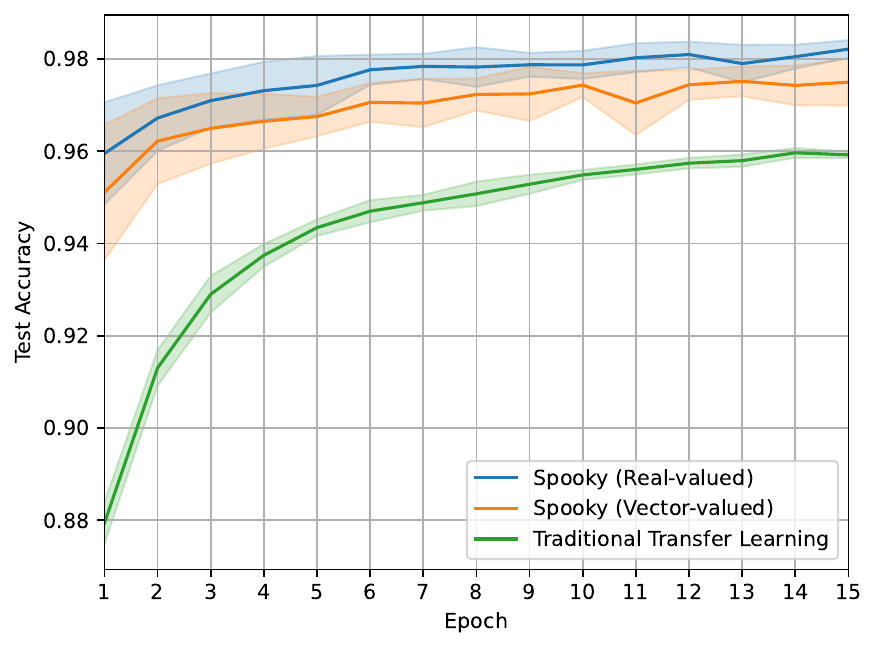}
    \caption{Test set accuracy per epoch for the baseline EfficientNetV2 (B0) and both Spooky TL variants. The shadowed area indicates the standard deviation.}
    \label{fig:acc_per_epoch}
\end{figure}

% We load the available EfficientNetV2 (B0) implementation from the keras.applications with weights pre-trained on the ImageNet dataset. This EfficientNet backbone is followed by a dense (softmax) layer with 8 neurons for classification, thus representing the baseline model. Our proposed Spooky EfficientNet consists of the backbone in which
The baseline comparison used was the available EfficientNetV2 (B0) implementation from the keras.applications library with pre-trained weights on ImageNet. This EfficientNet backbone is followed by a dense (softmax) layer with 8 neurons for classification. For the hypercomplex-valued models, we used the quaternion algebra detailed in Example \ref{ex:quaternions_example} and applied spooky transfer learning to the first convolutional layer of the EfficientNet, a layer with 32 kernels of size $3\times 3$ and stride 2. The resulting equivalent spooky layer produces an output with 128 feature channels, which is reduced to the original number of features using two methods: a traditional real-valued convolutional layer with 32 $1\times 1$ kernels (indicated by ``R'') and a quaternion-valued convolutional layer with 8 $1\times 1$ kernels (indicated by ``V''). All models were trained for 15 epochs using the Adam optimizer, cross-entropy loss, and batches of size 32. Table \ref{tab:acc_bloodmnist} shows the total number of trainable parameters and final test set accuracy per model, while Figure \ref{fig:acc_per_epoch} depicts accuracy per training epoch.

\begin{table}
    \centering
    \caption{Traditional vs spooky transfer learning trainable parameters and test set accuracy on BloodMNIST \cite{medmnistv2}.}
    \label{tab:acc_bloodmnist}
    \begin{tabular}{|l|c|c|} \hline
         & Trainable params. & Accuracy (\%) \\ \hline 
         EfficientNetV2 \cite{tan2021efficientnetV2} & $10,248$ & $95.92 \pm {\scriptstyle 0.07}$ \\ \hline 
         Spooky TL (R) & $14,376$ & $98.21\pm {\scriptstyle 0.19}$  \\ \hline 
         Spooky TL (V) & $11,304$ & $97.49\pm {\scriptstyle 0.51}$ \\ \hline 
    \end{tabular}
\end{table}

The results show notable gains in accuracy of the spooky transfer learning-based models over the traditional transfer learning model. Moreover, the added dimension reduction layers contain a small number of trainable parameters, $4128$ and $1056$ for the `R' and `V' variants respectively, thus representing a computationally efficient framework for recombination of features.

\section{Concluding Remarks} 
\label{sec:conclusion}

This paper formalized the concept of ghost features in hypercomplex-valued neural networks, illustrating how they naturally emerge when a traditional layer is embedded into a hypercomplex-valued one. Furthermore, we introduced spooky transfer learning, a novel methodology that embeds pre-trained real-valued layers into hypercomplex frameworks,  preserving the original features while inducing ghost features. Experimental results on the BloodMNIST dataset demonstrated the effectiveness of this approach, with spooky transfer learning surpassing traditional transfer learning on the EfficientNetV2 (B0) model. These findings confirm that ghost features provide redundant yet rich, discriminative information that enhances downstream task performance beyond standard real-valued representations. Furthermore, the proposed dimension reduction strategy ensures that these additional features are integrated into existing pipelines with high parameter efficiency. Ultimately, this work provides both a theoretical foundation and a practical path for leveraging pre-trained models and algebraic structures to design more robust deep learning models. A natural next step in this research is the exploration of information theoretic concepts, aimed at quantifying the contribution of ghost features, formally pinpointing its source, and further leveraging it to build lighter models by exploiting representational advantages.

\bibliographystyle{IEEEbib}
\bibliography{references}

\end{document}